*Research Paper*

# A Reinforcement Learning-based Adaptive Control Model for Future Street Planning

## An Algorithm and A Case Study

Qiming YE, Imperial College London, United Kingdom
Yuxiang FENG, Imperial College London, United Kingdom
Jing HAN, Tongji University, People's Republic of China
Marc STETTLER, Imperial College London, United Kingdom
Panagiotis ANGELOUDIS, Imperial College London, United Kingdom

**Abstract**

*With the emerging technologies in Intelligent Transportation System (ITS), the adaptive operation of road space is likely to be realised within decades. An intelligent street can learn and improve its decision-making on the right-of-way (ROW) for road users, liberating more active pedestrian space while maintaining traffic safety and efficiency. However, there is a lack of effective controlling techniques for these adaptive street infrastructures. To fill this gap in existing studies, we formulate this control problem as a Markov Game and develop a solution based on the multi-agent Deep Deterministic Policy Gradient (MADDPG) algorithm. The proposed model can dynamically assign ROW for sidewalks, autonomous vehicles (AVs) driving lanes and on-street parking areas in real-time. Integrated with the SUMO traffic simulator, this model was evaluated using the road network of the South Kensington District against three cases of divergent traffic conditions: pedestrian flow rates, AVs traffic flow rates and parking demands. Results reveal that our model can achieve an average reduction of 3.87% and 6.26% in street space assigned for on-street parking and vehicular operations. Combined with space gained by limiting the number of driving lanes, the average proportion of sidewalks to total widths of streets can significantly increase by 10.13%.*

## 1. Introduction

As the principal focus of city planning generally shifted from massive land development to localised urban renewal based on the built areas, developed cities are encouraging advanced technologies and pioneering practices to be deployed (Batty et al., 2012). This new wave of renovation depicts an intelligent, low-carbon, environmentally-friendly and human-centred urban space (Anthopoulos, 2017). Emerging technologies in the Intelligent Transport System (ITS) represent one strong incentive for urban renewal. The prominent disruptive technology would be autonomous vehicles (AVs). AVs transport is anticipated to be a safer and more efficient alternative to human-driven vehicles in urban mobility (Koopman and Wagner, 2017). Zhang (2015) modelled the shared AVs (SAVs) and found that they are likely to liberate over 80% of the existing parking space and fundamentally change the parking patterns since AVs could deliver the working class to downtown districts and autopilot to suburban districts for parking and charging. Meanwhile, Massar et al. (2021) and Rafael et al. (2020) concluded that electricity-powered AVs (EAVs) could reduce greenhouse gases (GHG) and air pollutants emission by 35% and 30%, respectively. However, a significant body of research, such as Duarte and Ratti (2018), predicted that the deployment of AVs to urban space would induce travel demands, compress active modes used for short-range trips, and exacerbate traffic conditions in already congested areas.

Undoubtedly, proactively upgrading the road infrastructure and installing advanced road facilities are the premises to adapt the future road space to AVs mobility. However, discussion on the future roads is on the rise, whereas concrete controlling techniques still represent a gap. Therefore, we propose a model that dynamically assigns ROW for sidewalks, AVs driving lanes and on-street parking areas in real-time to address these omissions. This control problem is formulated as a Markov Game and solved using a Reinforcement Learning (RL) method, namely the multi-agent Deep Deterministic Policy Gradient (MADDPG) algorithm. Our model has been applied to the road network of the South Kensington District and tested under three traffic conditions.

The contribution of our paper is both theoretical and methodological.

- Our method provides a novel solution to improve road traffic management of basic road geometries and complex road networks.

- This method can optimise the right-of-way (ROW) assignment tasks for sidewalks, driving lanes and on-street parking in traffic engineering and urban planning.

- As a building block to ITS and to reshaping the road space, our devised methods can further integrate with other ITS technologies.

The synopsis of the remainder of this paper is specified as follows —section 2 reviews the future road space, on-street parking operation and reinforcement learning. We specify our model and the proposed MADDPG algorithm in Section 3. Section 4 introduces the study case and traffic conditions for testing. Section 5 demonstrates results, and Section 6 discuss and concludes the paper.

## 2. Literature Review

### 2.1. Future Road Space and Complete Street Scheme

The future road space and how emerging technologies may reshape it are gaining popularity not solely confined to the arena of transportation, but urban planning, computer science and telecommunication. A growing body of studies has proposed, devised or improved critical technologies implemented to road space. For instance, adaptive Traffic Signal Control (TSC) schemes have been designed at intersections and crossings to enhance the throughput of fleets (Chu et al., 2019; Wang et al., 2020). From the perspective of urban planning and spatial design, inclusiveness, accessibility, user-friendly, and just right-of-way (ROW) are the principal qualities of future roads. Studies by Middel et al. (2019), Cadamuro et al. (2019) and Liu et al. (2019) applied Deep Learning (DL) techniques to measure walkability, connectivity, greenery and other qualities of the street space. Equipped with advanced critical techniques from ITS, such as Roadside Units (RSUs) and fast-charging facilities, the future road space is likely to realise a wide range of automated scenes that used to be described in science fiction. Barrachina et al. (2013) investigated the impact of the density and location for RSUs deployment on supporting Vehicle Communication (VC). In a similar setting, Anastasiadis et al. (2019) solved the location problem for electronic and fast-charging facilities for a fleet of auto-taxi in Chicago.

The most recognisable paradigm for street design is the complete street scheme, which mandates fixed proportions of road space for various road users (Smith and Baker, 2010). In practice, Dumbaug and King (2018) concluded that a generic ROW might comprise driving lanes, medians, sidewalks, cycle lanes and belts for facilities and urban furniture. Thus, spatially fixed ROW is provided to distinctive types of road users, and the right and safety of non-vehicular users are guaranteed. However, such schemes fall short of flexible arrangements over subdivision plans to address the changing traffic conditions (Karndacharuk et al., 2014).

### 2.3. On-Street Parking Operations

An adaptive control scheme is also expected to balance road space for on-street parking and driving, intelligently assign parking demands to the suitable locations at the right time. Allowing on-street parking can effectively reduce the distance of cruising to parking and shortens the last miles between Pick-Up and Drop-Off (PUDO) locations and destinations (Rodier et al., 2020). Jacobs (2016) stressed the function of on-street parking as a buffer between sidewalks and driving lanes, enhancing the sense of safety for pedestrians. Additionally, installing an on-street parking area could improve transport efficiency (Millard-Ball, 2019). As indicated by Biswas et al. (2017) and Marsden (2006), on-street parking saves space dedicated for parking, alleviating congestion by reducing cruising for parking.

As per our previous study, opening 50% of curb lanes for parking can effectively alleviate traffic delays by up to 27% (Ye et al., 2020). However, it also demonstrates a diminishing marginal benefit as parking provision rates or traffic saturation conditions change. Thus, a dynamic on-street parking provision scheme represents another essential function of the future road space.

### 2.4. Reinforcement Learning

Reinforcement Learning is one of the most significant Artificial Intelligence (AI) techniques, which addresses the optimal control problem through a trial-and-error approach (Sutton et al., 2018). Such RL controlling process can be simply understood as agent(s) learns the acting strategy with the environment by continuous interacting with the environment (Jaderberg et al., 2016). In recent years, RL has been widely deployed in critical fields of mathematics and robotics to solve real-world problems. Qiang and Zhongli (2011) summarised the applications of RL, including dispatching, robotic motions, Travel Salesman Problem (TSP) solving, and control strategies for ITS. For example, Chu et al. (2019) and Wang et al. (2020) applied RL to control traffic signals.

This process is also referred to as a Markov Decision Process (MDP), which is conventionally denoted by a tuple $< S, A, \mathbb{P}, R, \gamma >$ (Kaelbling et al., 1996; Van Otterlo and Wiering, 2012). By definition, $S$ represents the observed states of the environment. $A$ is the actions taken by agent(s). $\mathbb{P}$ indicates the transition probability of the current state $s$ to the next state $s'$. R is the reward, indicating the feedback from the environment. $\gamma$ parameterises the discount of future rewards. Given a multi-agent stochastic system, an MDP is formulated into a Markov Game (MG). Littman (1994) was among the earliest to introduce MG for RL-based problems. By definition, A MG can be expressed using a tuple $< N, S, A, \mathbb{P}, R, \gamma >$, adding the number of agents $N$ to the tuple of MDP.

## 3. Methodology

### 3.1. Problem Formulation

The identified research problem represents a sequential decision problem or a Markov Game. Concretely, given a road network of $G = (V, E)$, a directed edge $e \in E$ learns to decide the ROW proportions assigned to sidewalks ($\beta_{sidewalk}$), driving lanes ($\beta_{veh}$) and on-street parking operations ($\beta_{park}$) as illustrated in

Figure 1. The width of the edge and the proportion of the facility belt are denoted by w and $\beta_{faci}$. The objective is to maximise the reward R at each time slot $t \in T$ and feedback from the transport system for each edge $e$.

As expressed in Equation 1, this reward R comprises four sub rewards. As defined in Equation 2, $R_{sidewalk}^{t,e}$ denotes the proportional widths assigned for the sidewalk. Let NP and NV indicate the number of operational AVs and pedestrians in the system, and $p \in P$ and $n \in N$ denote a pedestrian and an AV, respectively. In Equation 3, $R_{ped}^{t,e}$ represents the pedestrian traffic efficiency of the focused edge. It first observes all AVs driving speed $v_{ped}^{t,e}$ and divides by the maximal allowed AVs speed $v_{ped}^*$, then calculates the expectation of such relative speed for all edges. In a similar setting, following Equation 4, $R_{ped}^{t,e}$ calculates the pedestrian traffic efficiency. In Equation 5, $R_{park}^{t,e}$ evaluates the effective service level of on-street parking operations, where $k_{dem}^{t,e}$ and $k_{park}^{t,e}$ indicate the edge-based on-street parking demands and parking provision. Constraint 6 enforces that all ROW parts sum to 1.

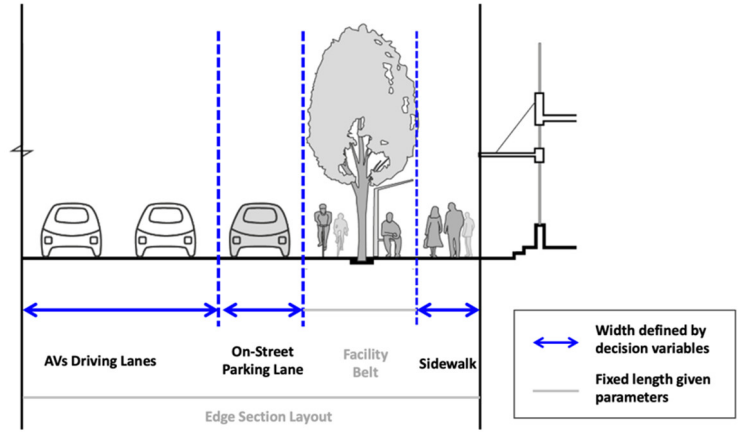

Figure 1. Illustration of the Right-of-Way (ROW) control problem.

$$maximise \quad R = \sum_{t=1}^{T}\sum_{e=1}^{E}(R_{sidewalk}^{t,e} + R_{ped}^{t,e} + R_{veh}^{t,e} + R_{park}^{t,e}) \qquad \forall e \in E, \forall t \in T \quad (1)$$

$$R_{sidewalk}^{t,e} = \beta_{sidewalk}^{t,e} \qquad (2)$$

$$R_{ped}^{t,e} = \frac{1}{NP}\sum_{p=1}^{P}\frac{v_{ped}^{t,e}}{v_{ped}^*} \qquad \forall p \in P \quad (3)$$

$$R_{veh}^{t,e} = \frac{1}{NV}\sum_{n=1}^{N}\frac{v_{veh}^{t,e}}{v_{ved}^*} \qquad \forall n \in N \quad (4)$$

$$R_{park}^{t,e} = \frac{k_{dem}^{t,e}}{k_{park}^{t,e}} \qquad (5)$$

$$s.t. \quad \beta_{sidewalk}^{t,e} + \beta_{veh}^{t,e} + \beta_{faci}^{t,e} + \beta_{park}^{t,e} \equiv 1 \qquad (6)$$

### 3.2. A Multi-agent Deep Deterministic Policy Gradient (MADDPG) Algorithm

We introduced a reinforcement learning method to address this sequential decision problem. Because that continuous action spaces define the decision variables $\beta_{sidewalk}$, $\beta_{veh}$ and $\beta_{park}$, we applied the Deep Deterministic Policy Gradient (DDPG) algorithm as the solution method. In addition, an equivalent number of $e$ Actor-Critic (AC) neural networks, in corresponding to $e$ edges, is architect to establish a multi-agent DDPG (MADDPG) algorithm for simultaneous and distributive training. The open traffic simulation platform SUMO facilitates the simulation of the transport system. Our MADDPG integrates with SUMO for updating the ROW configurations and retrieving the traffic states from the system.

## 4. Case Specification

### 4.1. South Kensington District

The experimental case for testing our proposed method is a 0.65 km$^2$ area at the heart of the South Kingston District in London. Figure 2 presents its general location and the principal functions of each land parcel. Land parcel from 1$^{st}$ to 8$^{th}$ now houses a wide range of essential heritages or culturally or artistically influential institutes. These include the Natural History Museum, the National Science Museum, the Victoria and Albert Museum, the Royal Albert Hall, the Royal College of Art, the Royal College of Music, the Royal Society of Geographical Society, and Imperial College London. Land parcels indexed from 9th to 18th represents low-density townhouses for residence, hospitality or consulate functions. The Exhibition Rd and Queen's Gate Rd are two main streets linking Hyde Park in the north and the two stations in the south.

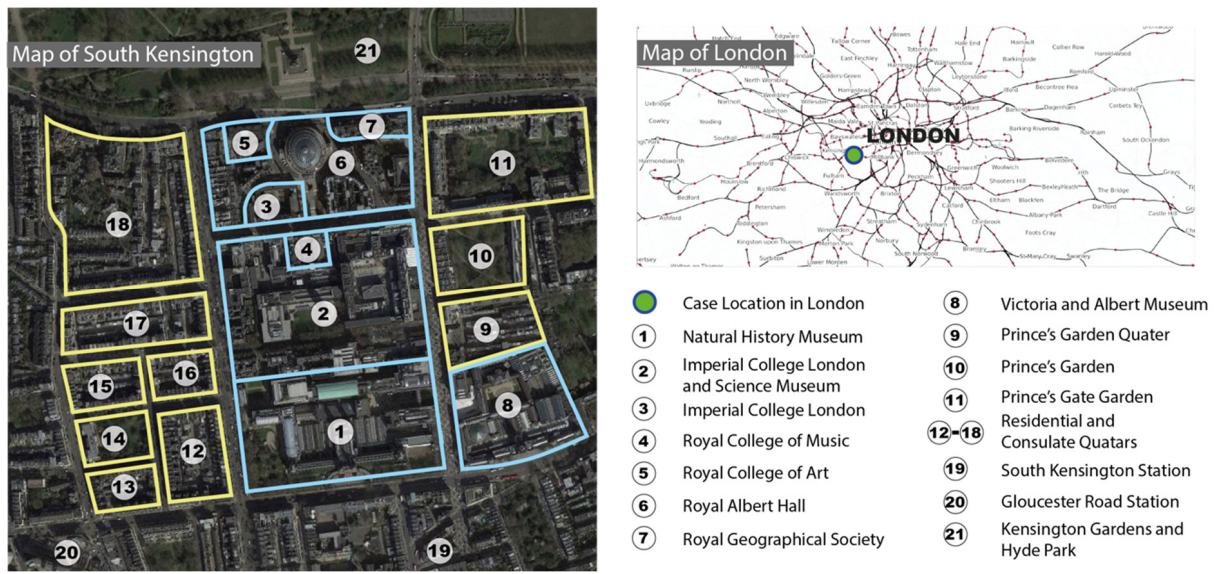

Figure 2. Location of South Kensington district and the principal functions land parcels of the study case. Base map sources: Google Map and OpenStreetMap

### 4.2. Local Traffic Patterns

Figure 3 demonstrates the concerned 58 edges and their current states. Traffic patterns in the heart region and periphery residential locations differ in travel demands of vehicles, pedestrians, and parking. On the one hand, these giant public facilities attract visitors or commuters via South Kensington Station, induce significant pedestrian travel demands along the Exhibition Rd in the peak hours. In addition, long queues in front of the museums and thousands of audiences instantaneously poured from the Royal Albert Hall, alarmingly increasing the sidewalk conditions.

Specific design strategies and traffic management policies have been deployed to prioritise pedestrians' ROW and address the on-street parking demands. First, in corresponding to the overwhelmingly crowded sidewalks of Exhibition Rd during peak hours, the Transport for London (TfL) department cooperated with the local council to redesign the road layout in 2012. Second, a shared space scheme was introduced upon the remaining driving lanes to grant active mobility with free access (Kaparias et al., 2013).

Meanwhile, in those residential locations, pedestrian walking and street activities have not been seriously considered. Their walking environment is not pedestrian-friendly nor aesthetically pleasant, as presented in Figure 3. The state-of-the-art policy allows on-street parking on either side of the streets due to low through traffic and high parking demands. For instance, at some locations, the street poles, greeneries, dustbins are arranged within a 1*m*-wide sidewalk space, forcing pedestrians to walk on drive lanes. The functionality of street life can be restored if we can smartly assign road space to appropriate usage.

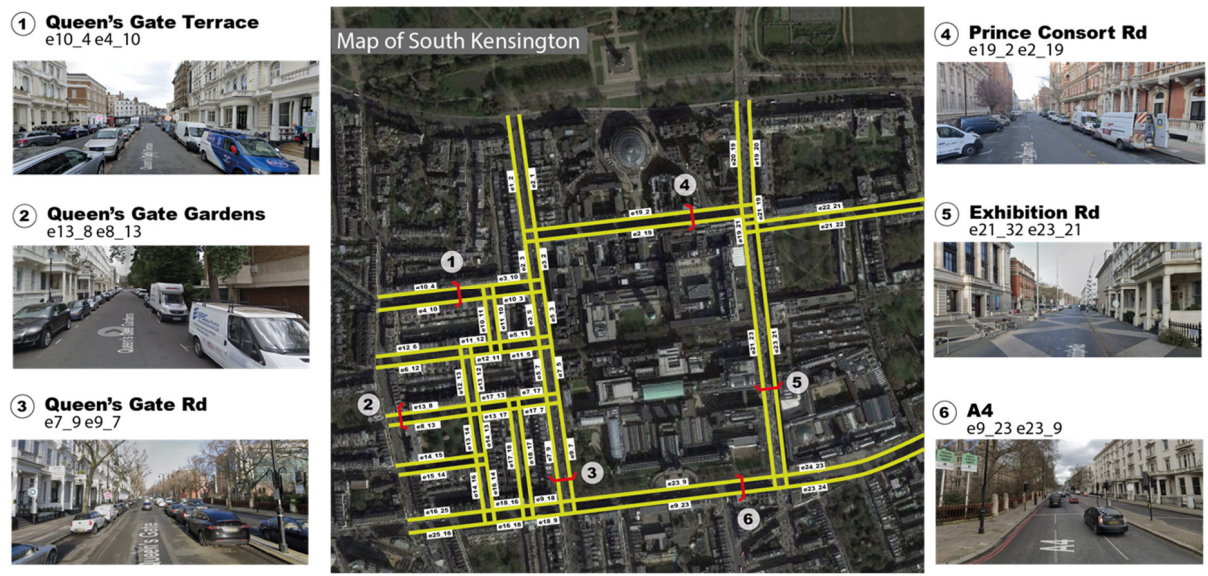

**Figure 3. Directed edges of the study case with street views of essential edges. Base map and street view images source: Google Map**

### 4.3. Traffic Patterns for Testing

Three traffic scenarios are synthesised for testing, which represents the low flow rate condition (Scenario 1), the high flow rate condition (Scenario 2) and pedestrian demands rich case (Scenario 3). In Scenario 1, This scheme could prioritise the width of sidewalks when through traffic and on-street parking demands are low. Scenario 2 represents a conflicting objective where flexible arrangements over the assignment of sidewalks, driving lanes and parking lanes should be considered. Finally, in Scenario 3, while keeping a high AVs traffic flow rate equivalent to Scenario 2, we randomly impose 1.5 to 2.5 times pedestrian travel demands per 30 minutes along the Exhibition edges.

## 5. Experimental Results and Discussion

Our proposed method has been tested against the South Kensington Rd network with three traffic scenarios. The MADDPG algorithm distributivity optimised the ROW plan for 58 directed edges (29 streets) within 150 training episodes. In this section, we present the principal experimental results, consisting of traffic simulation, spatial distributions of street layout improvements, and learning curves that indicate the optimisation performance of our proposed method.

### 5.1. Traffic Simulation Results

SUMO was employed to simulate AVs traffic, on-street parking operations of the fleet and pedestrian traffic. Its open API (TraCI) was used to retrieve simulation results at every 36 simulation seconds with predefined communication protocols. Each simulation step is defined by a fixed temporal length of 1,800 simulation seconds. Thus, a day is split into 48 slots. The three test scenarios represent the low flow rate case, the high flow rate case and pedestrian demands rich case.

Based on the travel demands encoded in the tested traffic scenarios, the peak flows of AVs in operation in the network are 63, 118 and 121, situating around 18:00 to 19:00. Peak pedestrian flows are 53, 102 and 169, respectively. One prominent problem of assigning travel demands into discrete time slots is the discontinuity of the transport environment regarding continuing effects of traffic delays and lane assignments. To solve this issue, at the end of each segment, those unfinished vehicular or pedestrian trips are stored using a data buffer externally and restored to continue their respective journeys at the beginning of the next simulation slot. In general, these trip restorations occurred 26, 38 and 41 times respectively for the simulation tasks of three scenarios. The driving features of AVs and physical characteristics of pedestrian movements are tabulated in Table 1 below.

Table 1. Physical parameters of pedestrians and AVs operations for traffic simulation in SUMO

| Type | Specifications | Values |
|---|---|---|
| | Minimal person following gap | 2.50 m |
| Pedestrian Movements | Maximum walking speed | 1.20 m/s |
| | Width of body | 1.00 m |
| | Acceleration | 2.60 m/s$^2$ |
| AVs Operations | deceleration | 4.50 m/s$^2$ |
| | Maximum driving speed | 13.00 m/s |
| | Time headway | 0.60 s |
| | Imperfection coefficient | 0.00 |
| | Speeds factor | 0.05 |
| | Length of vehicle | 4.00 m |

On-street parking operations are one of the critical simulation tasks of our experiment. Initially, each AV's decision of whether to park on a given edge follows a sigmoid function. Thus, the theoretical levels of on-street parking demands depend on the throughput of operational AVs in the system. However, according to the policy of prioritising pedestrian movements and sidewalk activities, on-street parking provision is limited to a rational level. A parking request expires if the parking provision on any edge along the planned route is insufficient.

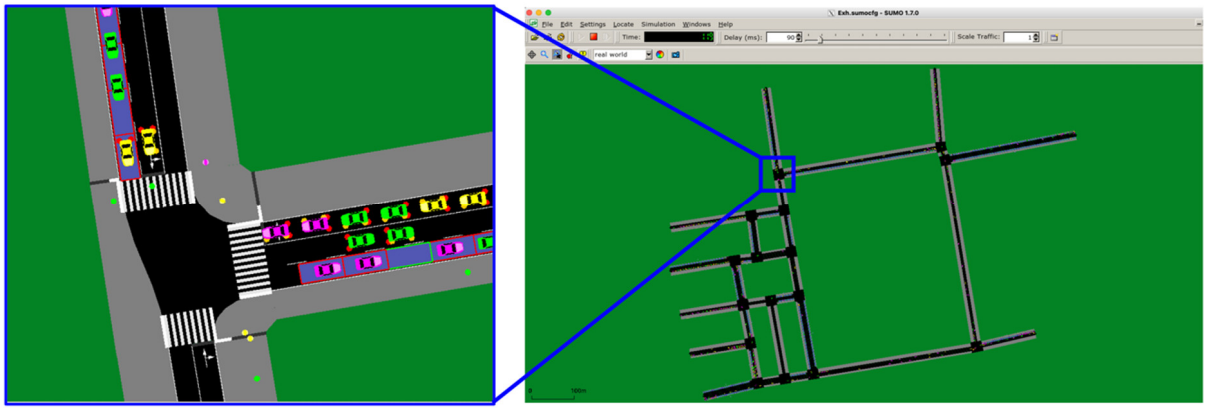

**Figure 4. Visualisation of traffic simulation of an arbitrary time slot using SUMO-GUI**

## 5.2. Changes in Traffic Efficiency and On-street Parking Operations

As our MADDPG algorithm learns to improve its ROW assignment strategies based on the flow of road users and on-street parking operations, the landscape of edge-based performance on efficiency and on-street parking provisions can significantly change. We quantified the varied performance, calculated their average value of every 50 episodes, and compared the early training episodes (Ep.0 – Ep.49) and the late training episodes (Ep.100 – Ep.149). Afterwards, we mapped the changes using a three-categorical demonstration to indicate the respective states of performance. In the following figures, we use green, red, and blue to highlight the changes, while different magnitudes of line width indicate the relative scales of difference to the mean level of changes.

Figure 5 presents the edge-based changes in pedestrian traffic efficiency. The results show that 13 edges (22.4%) have seen an increase in their pedestrian speed. The average reward of increased pedestrian efficiency is 3.94, and the maximum surge of 13.81 comes from edge 'e12_13'. Besides these improved samples, the number of edges whose pedestrian traffic efficiency has not changed is 44 (75.9%), counting for the majority of all three categories. Only one edge (1.7%) reported a loss in pedestrian traffic efficiency.

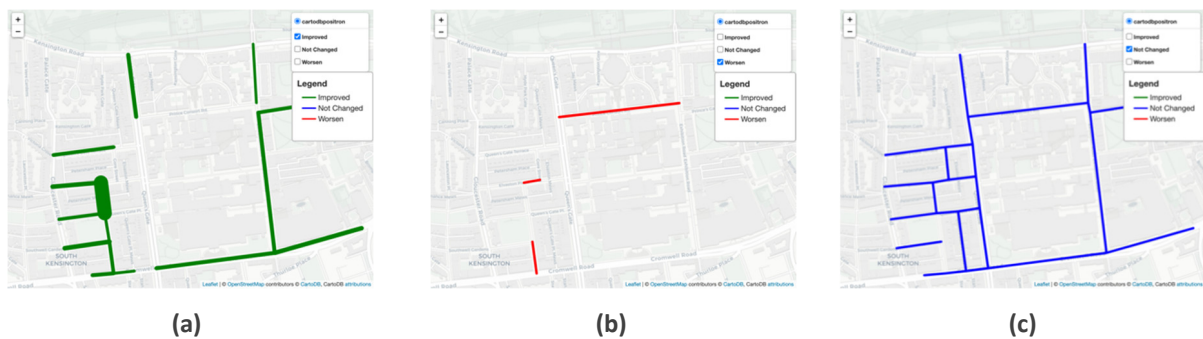

(a)            (b)            (c)

**Figure 5. Comparison of pedestrian traffic efficiency between early phase episodes (Ep.0-Ep.49) and later phase episodes (Ep.100-Ep.149). (a) Edges of improved pedestrian traffic efficiency; (b) Edges of worsened pedestrian traffic efficiency; (c) Edges of pedestrian efficiency not changed.**

Figure 6 demonstrates the changes in AVs traffic efficiency. In contrast with the predominant trend of not changed displayed in the pedestrian traffic efficiency variation, the vehicular efficiency has shown a quite diverged tendency. In concrete, 34 edges (58.6%) have increased their relative AVs operational speed, with an average rise of 72.06. Simultaneously, 22 edges (37.9%) have signalled a drop in vehicular efficiency. The average decrease is 84.27. Only two edges have claimed a steady performance in AVs operation.

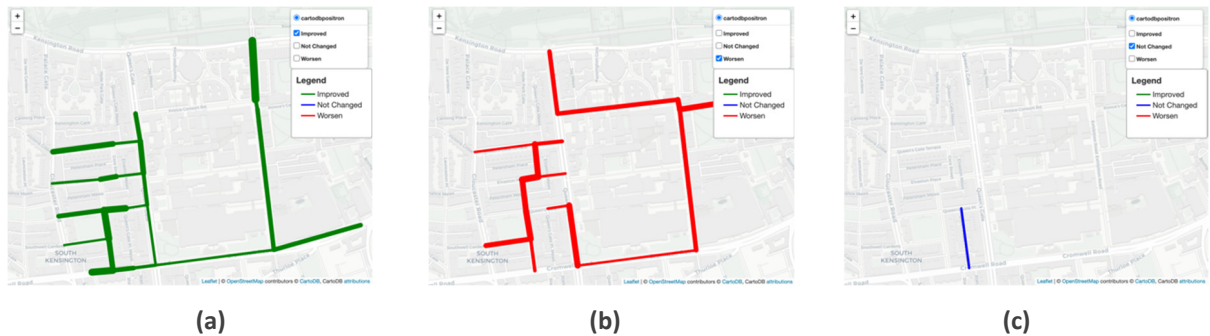

(a)            (b)            (c)

**Figure 6. Comparison of AVs traffic efficiency between early phase episodes (Ep.0-Ep.49) and later phase episodes (Ep.100-Ep.149). (a) Edges of improved AVs traffic efficiency; (b) Edges of worsened AVs traffic efficiency; (c) Edges of AVs efficiency not changed.**

Similarly, we have mapped the changes in on-street parking demands (Figure 7a), parking provision (Figure 7b), driving lanes provision (Figure 7c), and sidewalk provision (Figure 7d), respectively. Both parking demands and parking provision show a predominant trend of decrease. Specifically, 56 edges (96.6%) have experienced loss in parking demands at an average rate of -0.76 *pcu/edge*.

Regarding changes in ROW, street space assigned to on-street parking usage uniformly reduced in all 58 edges (100%), with which spatial proportion of on-street parking have dropped by 3.87% per edge. Similarly, all edges have decreased their spatial proportion assigned for vehicular operations by 6.26% per edge. On the contrary, our algorithm has increased the proportion of sidewalks among all 58 edges (100%), with an average increase of 91.3 in respective reward gains, namely +10.13% in the ratio of ROW assignment in equivalent.

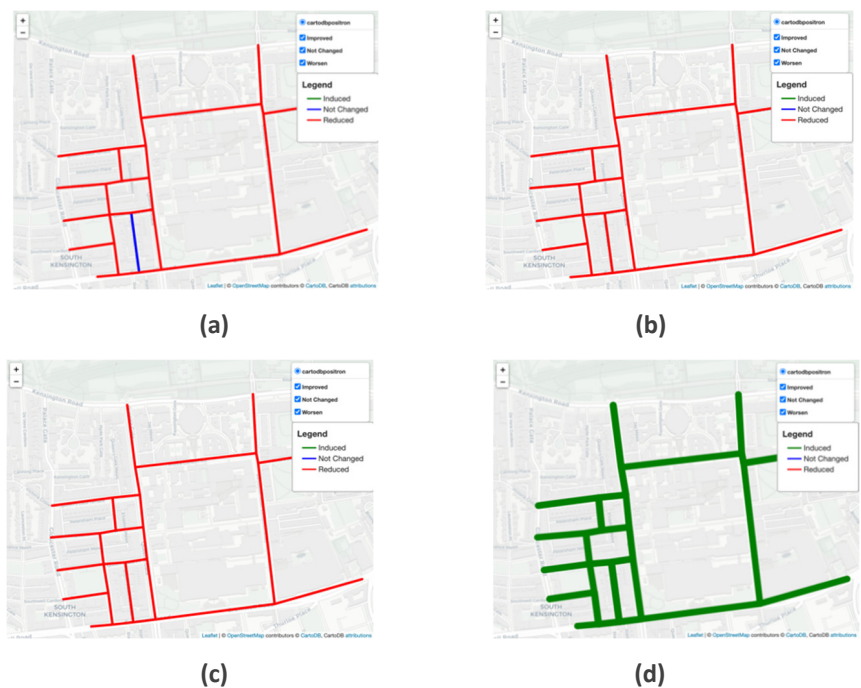

(a)            (b)

(c)            (d)

**Figure 7. Comparison of parking demands and ROW assignments between early phase episodes and later phase episodes. (a) On-street parking demands; (b) On-street parking provision; (c) Driving lanes provision; (d) Sidewalk provision.**

## 5.3. Learning Curves

To further demonstrate the effectiveness of our proposed method, we have examined the convergence feature of learning curves of the tested scenarios within the spectrum of 150 training episodes. We applied blue, green and red colours to distinguish the convergence patterns of respective cases visually. The convergence speed and searched optima are compared concerning the cumulative rewards (Figure 8a) and sub rewards (Figure 8b-8e). Finally, the convergence patterns of sub rewards are compared using Scenario 2 (Figure 8f).

The convergence patterns of the learning curves regarding the cumulative rewards in three cases demonstrate similar increasing trends. In terms of the initial cumulative rewards, Scenario 1 has the highest value of 2260.36, followed by 2189.94 of Scenario 3 and 2183.72 of Scenario 3. The average increases are 52.75, 215.16 and 244.45, reaching their perspective optima of 2313.12, 2379.97 and 2422.70. The episodic rewards of the low traffic flow scenario are 66.92 and 57.58 higher than those of Scenario 2 and 3. The amplitude of the learning curve generally narrowed with the optima approximates 2300 after Ep.90 for Scenario 1. Compared with a mild and regular increase pattern demonstrated by Scenario 1, the convergence patterns of Scenario 2 and 3 show several suddenly surge at Ep.28, Ep.63 and Ep.72, Ep.116 and Ep. 140. Such divergent patterns indicate a more robust optimal control with lower traffic flow conditions.

Regarding the sub rewards of AVs traffic efficiency, pedestrian efficiency, sidewalk proportions and on-street parking operations, the learning curves have shown distinct patterns in each scenario. In general, the sub rewards of Scenario 1 outperform that of the rest scenarios. First, the learning curves of AVs traffic efficiency shows similar patterns of the cumulative rewards. Such average sub rewards are 695.96, 648.40 and 680.47, respectively. Second, the sub rewards of Scenario 1 and 2 have shown similar patterns, as the mean rewards situate 985.25, 984.43, whereas the reward is 967.79 for Scenario 3. Third, rewards from evaluating the sidewalk proportions have shown uniformly increase in all cases, as the rewards raised from around 480.00 to approximately 610. In contrast, the rewards obtained from on-street parking operations significantly dropped from 79.81 to 8.73 on average. Noticeably, the changes in sidewalk reward and parking rewards strongly suggest the property of pro-sidewalk of our method. Figure 7f shows that among four sub rewards, the primary contribution comes from the pedestrian traffic efficiency, followed by AVs traffic efficiency, sidewalk and parking operations.

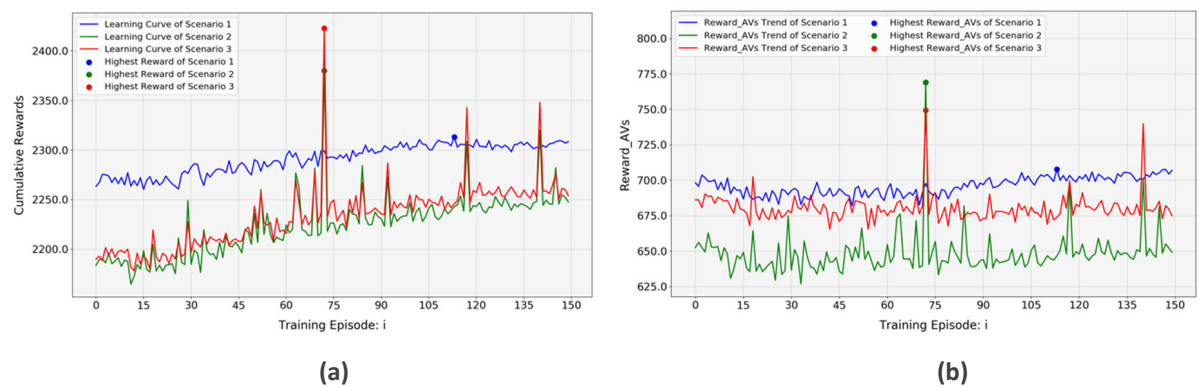

(a)      (b)

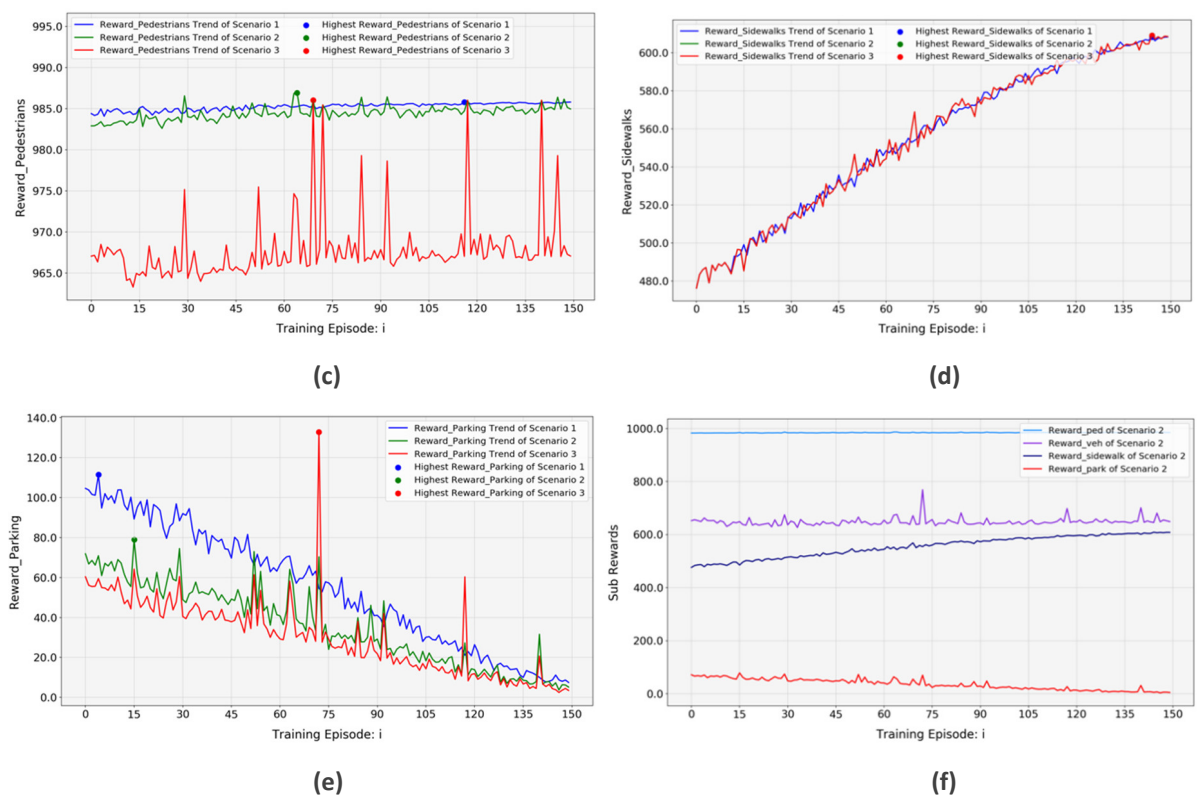

Figure 8. Learning curves of the tested scenarios. (a) Trends of cumulative rewards of three scenarios; (b) The learning curves of pedestrian traffic efficiency of three tested scenarios; (c) The learning curves of AVs traffic efficiency of three tested scenario; (d) The learning curves of sidewalk assignments of three tested scenario; (e) The learning curves of parking assignments of three tested scenario; (f) Comparison of learning curves of all four sub rewards of Scenario 2.

## 6. Conclusion

We present a reinforcement learning-based method to solve the ROW optimal control problem. The principal objective of the proposed method is to prioritise street usage as sidewalks. A MADDPG algorithm is implemented in the SUMO traffic simulator to optimise its ROW strategy in a trial-and-error approach. We have trained the method in 150 episodes against three cases of divergent traffic conditions: pedestrian flow rates, AVs traffic flow rates and parking demands. The proposed method has demonstrated convincing results in automatically calibrating its ROW layout as per varied flow conditions. This study could contribute to future road traffic management, parking management and street plan. It may also contribute to examining the implications of AVs mobility on the built environment and other road users.

Our method has considerably prioritised the ROW assignments to sidewalks while maintaining relative stability or increasing traffic efficiency. Despite reductions in the provision of on-street parking sites, the parking demands also reduce, liberating 3.87% space for pedestrian movements and other street activities. Combined with space gained by limiting the number of driving lanes, the average proportion of sidewalks to total widths of streets significantly raised by 10.13%.

The level of traffic flow rates also plays an inevitable factor in the optimal control of ROW. It was found that scenarios with lower traffic flow conditions always lead to higher rewards and faster convergence patterns. An increase in pedestrian flows can significantly affect pedestrian traffic efficiency, resulting in a limited influence on other sub rewards. This finding may indicate that even higher pedestrian capacity of the network can be achieved if vehicular efficiency or parking operations are of concern.

On-street parking operations can be effectively suppressed if limiting its provision. However, the induced street activities from the liberated sidewalks may also provide more short-period curb parking, or PUDO demands. Such latent effects will be examined in further studies. Other identified future directions include calibrating curb parking lanes to discrete and even angled parking sites and converting the facility belt into a new decision variable.